\documentclass[USenglish,oneside,twocolumn]{article}
\pdfoutput=1
\usepackage[utf8]{inputenc}
\usepackage[big]{dgruyter_NEW}
\DOI{foobar}
\usepackage{amsmath}    
\usepackage{graphicx}   
\usepackage{mathtools}  
\usepackage{multirow}   
\usepackage[caption=false]{subfig}
\usepackage{floatrow}
\usepackage{stackengine}
\usepackage{amssymb}
\usepackage{adjustbox}
\usepackage[hyphens]{url}

\DeclareMathOperator*{\argmax}{arg\,max}
\DeclareMathOperator*{\argmin}{arg\,min}
\DeclarePairedDelimiterX\set[1]\lbrace\rbrace{\setaux#1}\def\setaux#1|{#1\;\delimsize\vert\;}
\DeclarePairedDelimiterX{\norm}[1]{\lVert}{\rVert}{#1}

\cclogo{\includegraphics{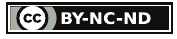}}

\begin{document}

\author[1]{Thomas Cilloni}
\author[2]{Wei Wang}
\author[3]{Charles Walter}
\author*[4]{Charles Fleming}

\affil[1]{University of Mississippi, E-mail: tcilloni@go.olemiss.edu}
\affil[2]{Xi'an Jiaotong-Liverpool University, E-mail: wei.wang03@xjtlu.edu.cn}
\affil[3]{University of Mississippi, E-mail: cwwalter@olemiss.edu}
\affil[4]{University of Mississippi, E-mail: fleming@olemiss.edu}

\title{\huge Ulixes: Facial Recognition Privacy with Adversarial Machine Learning}
\runningtitle{Ulixes: Facial Recognition Privacy with AML}


\begin{abstract}
{Facial recognition tools are becoming exceptionally accurate in identifying people from images. However, this comes at the cost of privacy for users of online services with photo management (e.g. social media platforms). Particularly troubling is the ability to leverage unsupervised learning to recognize faces even when the user has not labeled their images. 
In this paper we propose Ulixes, a strategy to generate visually non-invasive facial noise masks that yield adversarial examples, preventing the formation of identifiable user clusters in the embedding space of facial encoders. This is applicable even when a user is unmasked and labeled images are available online. We demonstrate the effectiveness of Ulixes by showing that various classification and clustering methods cannot reliably label the adversarial examples we generate. We also study the effects of Ulixes in various black-box settings and compare it to the current state of the art in adversarial machine learning. Finally, we challenge the effectiveness of Ulixes against adversarially trained models and show that it is robust to countermeasures.}
\end{abstract}

\journalname{Proceedings on Privacy Enhancing Technologies}
\DOI{10.2478/popets-2022-0008}

\startpage{148}
\received{2021-05-31}
\revised{2021-09-15}
\accepted{2021-09-16}

\journalyear{2022 }
\journalvolume{}
\journalissue{1}

\maketitle

\section{Introduction}

The rapid rise of the Information Age has left a gap in the legal infrastructure protecting users' privacy. Various jurisdictions around the world are upgrading their privacy-protection legislation, such as in Europe with the European Union General Data Protection Regulation \cite{gdpr}, but it is difficult to enforce such laws when data can be trivially transferred to other legal jurisdictions. Image data is particularly concerning because of its invasive applications and the lack of relevant legislation in existing systems \cite{LawsLack}. It is thus advisable for users of online services to protect themselves where other protection measures are not in place.

We focus on a common feature in many online services: photo tagging. Most consumer applications that provide services for the storage and distribution of images, such as Facebook and Flickr, make use of facial recognition tools \cite{PhotoTagging}. We propose a tool for users to protect their privacy by masquerading their face, called \textit{Ulixes} in honor of the legendary man who escaped death by stripping himself of his own personally identifiable information. 

There currently exist two systems whose goals and methods are similar: Fawkes \cite{Fawkes} and Face-Off \cite{face-off}. However, the first requires poisoning entire training datasets and is effective only when a person is disguised among a large number of other individuals, and the second introduces levels of noise that completely change the appearance of a face. Ulixes is employed in the more challenging case where an end-user is unable to poison datasets that may already contain some of his/her images and desires a high protection level with minimal visual changes.

Inspired by current adversarial machine learning techniques for classifiers, we design a new cloaking technique specifically for facial recognition systems that generate embeddings and explore various flavors of it. Ulixes can be employed on a number of different models and can decrease their accuracy by over $90\%$, without having to poison training datasets by adding an imperceptible amount of noise. In addition, Ulixes is highly efficient: a face image can be anonymized in less than 3 seconds on a low-end laptop without a GPU.

Our main contributions are as follows:
\begin{enumerate}
    \item We propose a novel analytical solution to creating adversarial examples built on the \textit{Triplet Loss} function common to many facial recognition systems today.
    \item Ulixes is able to change how facial recognition systems process images so drastically that it becomes impossible to perform either verification or identification tasks, either supervised or unsupervised.
    \item Ulixes is both more effective and significantly faster than any state-of-the-art system today. 
    \item We demonstrate the generalization of Ulixes to various models and an online API with a series of challenging transferability experiments.
\end{enumerate}

\section{Literature Review}
\noindent\textbf{Facial Recognition} Facial recognition is a branch of computer vision that studies techniques to extract information from face images. Typical tasks include face verification, a binary decision problem where two faces are compared to assess whether they represent the same person, and face identification, a classification problem that consists in assigning a label, or a name, to a face image.
By definition, facial identification is a classification task. Classification, however, allows only a fixed number of output classes in the network, which is quite impractical for facial recognition as a new network would have to be designed and trained every time a new person was to be identified. The solution is to transition from a classification task to a more regression-like task with networks that generate meaningful representations of faces in the form of numerical vectors.

One of the most accurate face recognition models today is FaceNet  \cite{FaceNet}, proposed by Google in 2015. It introduces a new lightweight strategy to identify faces: a deep CNN trained to optimize the representation of a face, which is described as a 128-dimensional numerical vector. Among the major contributions of FaceNet is the definition of a new metric for the cost function: the \textit{triplet loss}. Older models compute the loss by looking at the difference between the expected and the computed similarity between faces' embeddings. Examples include DeepFace, which uses Cross-Entropy Loss \cite{DeepFace_1} \cite{DeepFace_2}, and the series of DeepId, which use Cross-Entropy Loss alone or together with the Euclidean Loss \cite{DeepId_1} \cite{DeepId_2} \cite{DeepId_3} \cite{DeepId_4}. Triplet loss, instead, uses triplets of face images: an anchor $a$ with identity $ID$, a positive example $p$ also with identity $ID$, and a negative example $n$ with identity $ID' \neq ID$. The loss is calculated for each triplet as to how close the anchor is to the negative sample and how distant it is from the positive sample.
Eq. \eqref{eq:tripletloss} formalizes the triplet loss function $L$ of a model $f$ for a single triplet of images, where $m$ is how much greater the distance $a$ to $n$ should be than $a$ to $p$.
\begin{equation}
\label{eq:tripletloss}
L(a,p,n) = \max(\norm{f(a)-f(p)}_2 - \norm{f(a)-f(n)}_2 + m,\; 0)
\end{equation}
\noindent\textbf{Adversarial ML in Computer Vision} Adversarial Machine Learning is a relatively new area of study in artificial intelligence that looks at how neural networks can be fooled by feeding them deceptive inputs \cite{AdversarialML}. The techniques studied in adversarial machine learning are cyber attacks by nature and can be divided into two categories based on the resources available to the attacker: \textit{white-box} and \textit{black-box}.
When an adversary has access to the cost function of the victim model (or to the victim model as a whole), the attack is of a \textit{white-box} type. While the experiments proposed in this paper are carried out in a white-box setting, we will show how their results are transferable in a black-box fashion to other models we do not have internal access to. Black-box attacks are the most interesting because they are potentially effective against unknown victim models.

The \textit{Fast Gradient Sign Method} (FGSM) is one of the earliest examples of techniques in this field. It was proposed in 2014 in research by Goodfellow et al. \cite{AdversarialExamples} and applied to neural networks for the classification of images, such as in DeepFool  \cite{DeepFool}. The algorithm consists in computing the gradient of the loss function with respect to the input for a given model, and then images are altered in the direction opposite to the gradient in order to maximize the loss.

An improvement over FGSM is what is commonly referred to as \textit{One-step Target Class} \cite{AdversarialMLAtScale}. While FGSM simply decreases the confidence over the most likely target, the one-step target class method increases the confidence of a target class $y' \neq y$. This also decreases the confidence for $y$. 

Both the \textit{One-step Target Class} and the \textit{Fast Gradient Sign Method} are \textit{one-shot} methods: they are executed only once. It is possible, however, to repeat this alteration process with small alterations at each step to slowly bring an image to be misclassified. The \textit{Basic Iterative Method} is the iterative equivalent of the FGSM as the \textit{Iterative Least Likely Class} is to the \textit{One-step Target Class} \cite{AdversarialMLPhysicalWorld} \cite{BreakingDL}.

An important distinction is to be made between adversarial examples crafted in a \textit{targeted} or \textit{non-targeted} fashion. As described in \cite{targeted_untargeted} and seen in previous examples of adversarial machine learning techniques, \textit{non-targeted} attacks have the sole goal of producing adversarial examples that can hinder a neural network's accuracy. On the other hand, \textit{targeted} attacks have the purpose of bringing a neural network to classify an input as a specific class, called the target. Throughout this paper, both types of attacks are explored.

\noindent\textbf{Privacy-Preserving Adversarial ML} Shan et al. \cite{Fawkes} propose \textit{Fawkes}, a system to make faces in images unrecognizable by facial recognition tools. It does so by applying cloaks on face images computed by exploiting a feature extractor's output: generating a cloak is an optimization problem, where a cloak can only introduce a limited amount of noise (measured using the DSSIM score, see Section \ref{sec:algorithm}), while at the same time shifting the embedding of an image towards that of another person's image. Fawkes is effective in cloaking the identity of a person when its images are to be classified among many other people's images and an attacker has only access to cloaked images of said person. 

Ulixes is distinguishable from Fawkes because of two major advantages. First, in our threat model, the attacker uses a classifier trained on clear or cloaked images, whereas Fawkes requires a complete dataset poisoning. When even just a third of the training dataset is uncloaked, its effectiveness is cut almost in half. Second, Ulixes is computationally more efficient: it is $20 \times$ faster than the fastest of Fawkes' operating modes, and $200 \times$ faster than its standard operating mode.

A concurrent conference submission to Fawkes is FoggySight \cite{FoggySight}, which explores a very similar problem. While the threat model and the adversarial example generation methods of the two are quite similar, FoggySight uses pre-trained models to compute cloaks, which is a superior strategy to using models that need to be retrained. Additionally, FoggySight follows a different, intriguing concept: users of the system are seen as "protectors" of online communities, as they upload cloaked images on the Internet as decoy so that other users can have their privacy protected. Once a large number of cloaked images is online, models trained on newly formed datasets may not produce accurate results.

LowKey \cite{LowKey} closely follows Fawkes and FoggySight in terms of system design, threat model, and purpose. By poisoning the reference dataset with adversarial examples generated by cloaking faces, LowKey can render facial identification systems ineffective. Differently from previous works, however, LowKey includes a perceptual metric score in the loss function used to compute cloaks and uses an ensemble of models and blurred images to make its cloaks more robust.

Another related work is \textit{Face-Off} \cite{face-off}, which explores a wide range of innovative loss metrics to generate adversarial examples against face classifiers. It is unclear whether its attacks are directed at networks trained with cloaked or uncloaked images, but since it is not specified we assume the latter to be the case, thus providing a major improvement over Fawkes. Face-Off achieves a substantial relative accuracy decrease (about $-50\%$) on three commercially available face recognition APIs, however at the cost of very visually intrusive noise distortions and long processing times. Ulixes differs from Face-Off in that we explore a number of attack models that exploit the same loss function that the victim classifier was trained with, and show the efficacy of our attacks in supervised and unsupervised classification tasks too, on top of verification. Additionally, Ulixes' cloaks are almost imperceptible and can be computed in less than 3 seconds on a low-end CPU.

Recent research \cite{datapoisoning} shows how protection against facial recognition systems with data poisoning strategies can give a false sense of security. In the authors' investigation, both Fawkes and LowKey are unable to successfully anonymize images if an attacker waits for future developments in facial recognition technologies. These developments include training more robust models to resist the images' perturbations and detectors of data poisoning attacks. Such new insights further show the need for a non-data poisoning-based cloaking strategy for online data anonymization.

\section{System Design}

The entire process of facial recognition can be summarized as follows: localization of the face in an image, isolation and (optional) alignment and pre-processing of the face, generation of the embedding of the image (feature extraction), and comparison of the embedding against another one (for verification) or many other known embeddings (for identification). 
The goal of the cloaking  system is then to apply a very small alteration to a face in an image so that the outcome of the last step is erroneous.

\subsection{Threat Model}
We assume an \textit{attacker} to be an entity interested in performing accurate face recognition and having available a large dataset of labeled images to use as training. This training dataset can include samples of people whose images are later to be cloaked; additionally, its content is unknown and it cannot be poisoned. The attacker also has infinite computing power and performs facial recognition on only a small pool of labeled or unlabeled samples, thus making it easier to predict the correct class.

The \textit{defender} on the other hand is a single person interested in maintaining privacy. Their computing power is rather limited (on the scale of a consumer laptop) and so is their time (for instance, a user cannot wait as long as 5 minutes for a single image to be anonymous before uploading it on a social media, when the time required to shoot that image is likely a few seconds). The defender also does not know what facial recognition tools an attacker may use, but it has white-box access to one (or several) potential models.

\subsection{Cloaking Algorithm}\label{sec:algorithm}

Face images, whose pixels are each made of three values (one per RGB channel), are matrices $x$ in the input space defined by 
$X = \{x \in [-1,1]^{H \times W \times C}\}$, where $C=3$ is the number of channels and $H$ and $W$  are the height and width of the image, respectively. An encoder is a neural network $f: X \to E$ that maps an input image $x$ onto a vector $e \in E$ called an embedding. $E$ is a subspace of $\mathbb{R}^D$ which varies depending on the encoder used. The encoder is trained so that the Euclidean distance between two embeddings is a measure of the similarity (or dissimilarity) of the related faces. 

We then make the following hypothesis: cloaked image can be generated by adding a noise mask $\epsilon$ such that its resulting embedding is distant to its original one. In determining how far away is enough for a face image to be anonymized we use a checker $\phi$:
\begin{gather}
\phi(a, p, n) =
\begin{aligned}
\begin{cases}
  0, & \text{if}\ d(a, p) \leq d(a, n) \\
  1, & \text{otherwise}
\end{cases}
\end{aligned}\nonumber
\shortintertext{where}
d(x_1, x_2) = \norm{f(x_1)-f(x_2)}_2
\label{eq:phi}
\end{gather}

Where $a$ is the image to anonymize, and $p$ and $n$ are the positive and negative examples respectively. When $a$'s embedding is closer to $n$'s than to $p$'s, the image is said to be anonymized.

The core difficulty is how to find a noise mask $\epsilon$ that can effectively shift the embedding of a face away from its original counterpart and also be visually non-intrusive. It is necessary to keep the noise mask as small as possible: placing a black oval on top of a face can surely cloak it, but the resulting face would also be unrecognizable by any human being. Previous works \cite{Fawkes} solve this with a minimization problem, in which a dissimilarity metric, the DSSIM score \cite{DSSIM}, is used to measure how dissimilar two images are. Creating a noise mask $\epsilon$ is defined as the minimization problem of finding a value of $\epsilon$ that causes a face to be mistakenly classified and keeps the DSSIM to a minimum.

The triplet loss function gives a measure of how close the anchor is to the positive and negative example, and we use the one that is part of FaceNet \cite{FaceNet} to carry out most of the experiments in this paper. By inverting the positive with the negative example, the triplet loss shows how similar (or close) the anchor is to the negative example, and how far it is to the positive one. Its gradient with respect to the input to the network defines how an image should be "updated" in order to make its embedding farther away from the positive example and closer to the negative one: we refer to this as the adversarial triplet loss function, or $L_{ADV} (a,p,n)$ as (see Eq. \eqref{eq:tripletloss} for a definition of $L$):
\begin{equation}
\label{eq:l_adv}
L_{ADV} (a,p,n) = L (a,n,p)
\end{equation}

The problem can  be formalized as crafting an adversarial example $x'=x+\epsilon$ as in Eq. \eqref{eq:minimization_problem}, where $\hat{x}$ is a negative example for $x$.
\begin{gather}
\argmin_\epsilon \left\{ DSSIM(x, x+\epsilon) \right\}
\nonumber \\
\shortintertext{subject to}
x+\epsilon \in [-1,1]^{W \times H \times C} \;\;\text{and}\;\; \phi(x+\epsilon, x, \hat{x}) = 1
\label{eq:minimization_problem}
\end{gather}


Given the gradient vector of the adversarial loss function $\nabla_x L_{ADV}$, we scale it so that its maximum norm is $0.01$ as in Eq. \eqref{eq:GradientNormalization}, thus resulting in a small perturbation. As pixel color values are in the range of $[-1,1]$, an additive noise of $\pm 0.01$ corresponds to a $0.5\%$ alteration of a channel value, or $0.15\%$ if pixels are considered as a whole. 
Formalized, the problem is to iteratively look for a cloak $\epsilon$ to render an image $x$ adversarial as $x'_N=x+\epsilon_N$, where the number of cloaks $N$ is unconstrained:
\begin{gather}
\epsilon_N = \sum_{n=1}^N scale \left( \nabla_x L_{ADV}(x+\epsilon_{n-1}, x, \hat{x}) \right), \quad \epsilon_0 \to 0
\nonumber
\shortintertext{where}
scale(v) = \frac{v} {\norm{v}_{\infty}} \times 0.01
\label{eq:GradientNormalization}
\end{gather}

Given this definition of the algorithm, the remaining problem is to find optimal choices for the positive and negative examples, $a$ and $n$.  This is a question that we explore in Section \ref{sec:exploratory}.

\subsection{Components}

\noindent\textbf{Early Stop Condition}\\
The hypothesis that a face image is anonymized once its new embedding is significantly different from its original one holds true when an identifier has only one sample from the same person to compare a face against. That is if there is only one face image $f_1$ in a labeled database belonging to person $p$, then if a new image $f_N$ of the same person is cloaked with $\epsilon$ so that the L2 norm between the embeddings of $f_1$ and $f_N+\epsilon$ is large enough, then there is no way to associate $f_1$ with $f_N$ and $f_N$ is said to be cloaked.

If there are multiple images in a database belonging to the  person who is trying to cloak a new face image $f_N$, such as $f_1$ and $f_2$, the situation is more complicated. Applying a noise mask on $f_N$ to make it dissimilar to $f_1$ may not have the same effect on the distance from $f_N$ to $f_2$. It is possible that a cloak $\epsilon_1$ that makes $f'_N = f_N + \epsilon_1$ dissimilar from $f_1$ and another cloak $\epsilon_2$ that makes $f''_N = f'_N + \epsilon_2$ dissimilar from $f_2$ cancel out each other, e.g. $\epsilon_1 \approx - \epsilon_2$.

It is thus necessary to prevent the cloaking algorithm from falling into a loop of generating noise masks effective for a particular data sample but not for another. This is solved by picking an early stop condition independent from the data samples being used: cloaks $\epsilon_k$ are generated for as long as the $L_\infty$ norm of the loss function is positive and the cloaks have a significant impact on the anonymization of a face. Such significance is measured as the change in the distance from the anchor to the positive example across iterations, and it must be greater than a threshold $t$. Finally, the DSSIM score between the original and the cloaked image is used to set an upper bound on the amount of noise that can be added to an image.

The original FaceNet paper argues that two faces are said to belong to different people when their embeddings' L2 distance is greater than $1.242$. Therefore there must be a minimum distance between a cloaked image and its original version to say that the two images belong to different people. This is achieved by tuning the margin used in the triplet loss function, which defines how different two distances must be to consider a triplet to be good (see Eq. \eqref{eq:tripletloss}). The maximum distance between two embeddings is limited by the mathematical conditions imposed by the hyperspace of the model's output and the constraints on the embeddings. In FaceNet, embeddings can be at a maximum distance of $4$ units, and only in extreme cases. In our experiments, we registered maximum distances around $1.8$, which means that it may not be possible to push two embeddings farther away than that, considering the high dimensionality of the feature space and the requirement for noise masks to be minimal. Consequently, we carry out experiments with margin values between $0.2$ and $2$. Varying this value has an effect on the intensity of noise masks introduced.

\begin{figure}
\centering
\includegraphics [width=0.6\columnwidth] {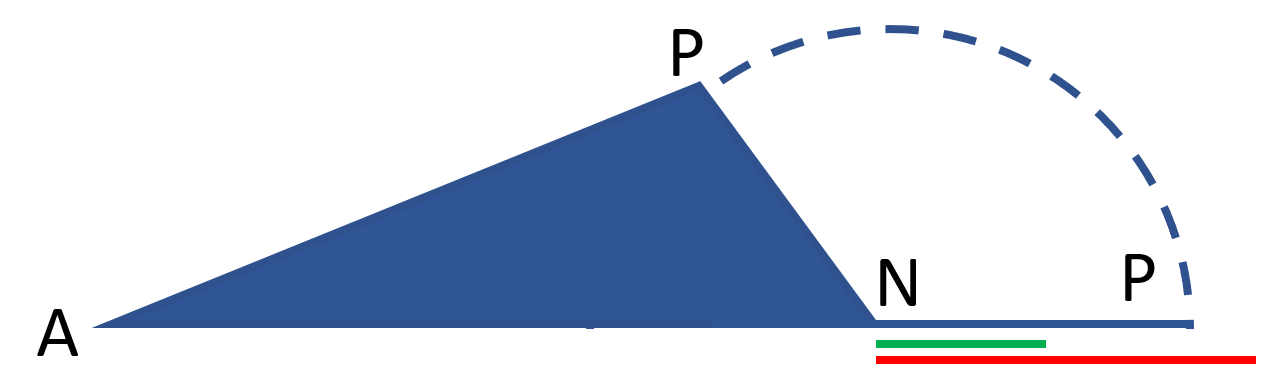}
\caption{\label{img:triangle}
    Distance requirement between positive and negative example relative to the margin. If the margin is small, such as that in green, then $\overline{NP}$ can be small. If it is larger though, such as that in red, $\overline{NP}$ must be larger than it or the loss cannot converge to $0$.
}
\end{figure}

\label{par:ExampleSelectionAndMargin}
In addition to these considerations, it is also possible that the margin is set to be too large for a particular image. 
Given a triangle formed with vertices $P$, $A$, and $N$, as in Fig. \ref{img:triangle}, the triplet loss function would return $0$ once $\overline{AN}+m < \overline{AP}$ is satisfied. By the triangle inequality this is possible only if $\overline{NP} > m$, which means that if the distance between the negative and positive examples is less than the margin $m$, the loss function will never return $0$ and the algorithm iterates indefinitely. This is more reason to set a threshold $\Delta$ distance, and we determined experimentally that the threshold $t=0.01$ leaves enough room for the cloaking algorithm to be effective.

\noindent\textbf{Example Selection}\\
FaceNet is a DNN trained to extract features from individual face images. The loss function, however, requires three images: an anchor, that is the image to cloak, a positive example from the same person, and a negative example of another person. Alternatively, the positive and negative examples can be the most similar and the most dissimilar embeddings' images.

In our experiments, at the first iteration, the positive example is always equal to the anchor. This significantly eases the example selection process and increases consistency among experiments. Deciding a positive example in any other way would not be meaningful: the closest example from a dataset may or may not belong to the same person, so it would introduce inconsistencies across experiments. This method does not impact the model negatively: starting from the second iteration, the anchor is different from the positive example (because of the cloak added). As for the negative example selection, we propose five different options, as explained in the following section.

\subsection{Dataset}

We use a small testing dataset to compare the effectiveness of different cloaking system configurations. This is made of 10 images  from each of 25 randomly selected identities from the FaceScrub \cite{FaceScrub} dataset. A dataset of this size would be too small for training a model, but it is sufficient when used only to compare the efficacy and performance of the algorithm against a pre-trained model when adjusting its parameters.

\noindent\textbf{Pre-processing Pipeline}\\
Images in all experiments included in this paper are pre-processed to conform to the requirements of the models they are processed by. The list of steps followed for each image is:
\begin{enumerate}
    \item Given a person's image read as an integer matrix in the range $[0,255]$, the first step is localizing the boundaries of the face. This is done using MTCNN \cite{mtcnn}. The face is then cropped out.
    \item Depending on the model used, the image is then resized to the required shape. In the specific case of Facenet, images have a shape of $160\times160\times3$.
    \item Depending on the model used, the image is normalized according to the requirements. In the specific case of Facenet\footnote{\url{https://github.com/davidsandberg/facenet/blob/master/src/facenet.py}}, images are normalized by subtracting $127.5$ from all color channel values, and then dividing by $128$: the output is, therefore, a matrix of values in range $[-1,1]$.
\end{enumerate}

\section{Exploratory Experiments}
\label{sec:exploratory}
\label{par:GradientNormalization}
In this section, we propose a series of experiments that explore different ways to choose a target negative example for the generation of cloaks. We use \textit{triplet loss} in all experiments due to its wide use in facial recognition systems today \cite{FaceNet} \cite{openface} \cite{triplet_loss_1} \cite{triplet_loss_2} \cite{triplet_loss_3} \cite{triplet_loss_4}. The experiments included in this and later sections are performed using a low-end laptop featuring an Intel Core i5-8250U processor, no GPU, and a vented cooling system.

\noindent\textbf{Relevant Data}\\
We evaluate the efficacy of each algorithm for different margin values through the \textit{verification} task, using the testing dataset described earlier. For each face in the set, there are $9$ other faces with the same identity, and $240$ faces with different identities. Consequently, a total of $9\times250 / 2 = 1,125$ distinct matching pairs and $240\times250 / 2 = 30,000$ distinct mismatching pairs of images are available. We look at both lists of pairs and calculate the related \textit{True Positive} and \textit{False Positive} rates.

We then measure visual impact on the images using the \textit{Structural Dissimilarity} score (DSSIM) \cite{DSSIM} between pairs of original and cloaked images. Each of the $250$ sample images used is compared against their cloaked counterparts to calculate the DSSIM score. All DSSIM results are then analyzed statistically.

\subsection{Method 1 - Farthest Sample}

The farthest face embedding is the one that should theoretically be most dissimilar to the face which is being cloaked and intuitively is the most appealing choice as a negative example. This method is similar to what is used in Fawkes \cite{Fawkes}, and it is an adaptation of the \textit{Least Likely Class} method proposed in  \cite{AdversarialExamples}.

\noindent\textbf{Algorithm}\\
The face whose embedding is farthest from that of the image being cloaked is picked as a negative example $\hat{x}$ from the pool of available images $D$:
$$\hat{x} = \argmax_{\hat{x} \in D} \left\{\norm{f(x)-f(\hat{x})}_2\right\}$$
A triplet is then formed using the face to cloak as both anchor $a$ and positive example $p$, and the selected embedding's face as the negative example $n$.

\begin{figure}%
\centering
\hspace{-0.5cm}
\sidesubfloat[]{{
    \frame{\includegraphics[width=0.3\columnwidth]{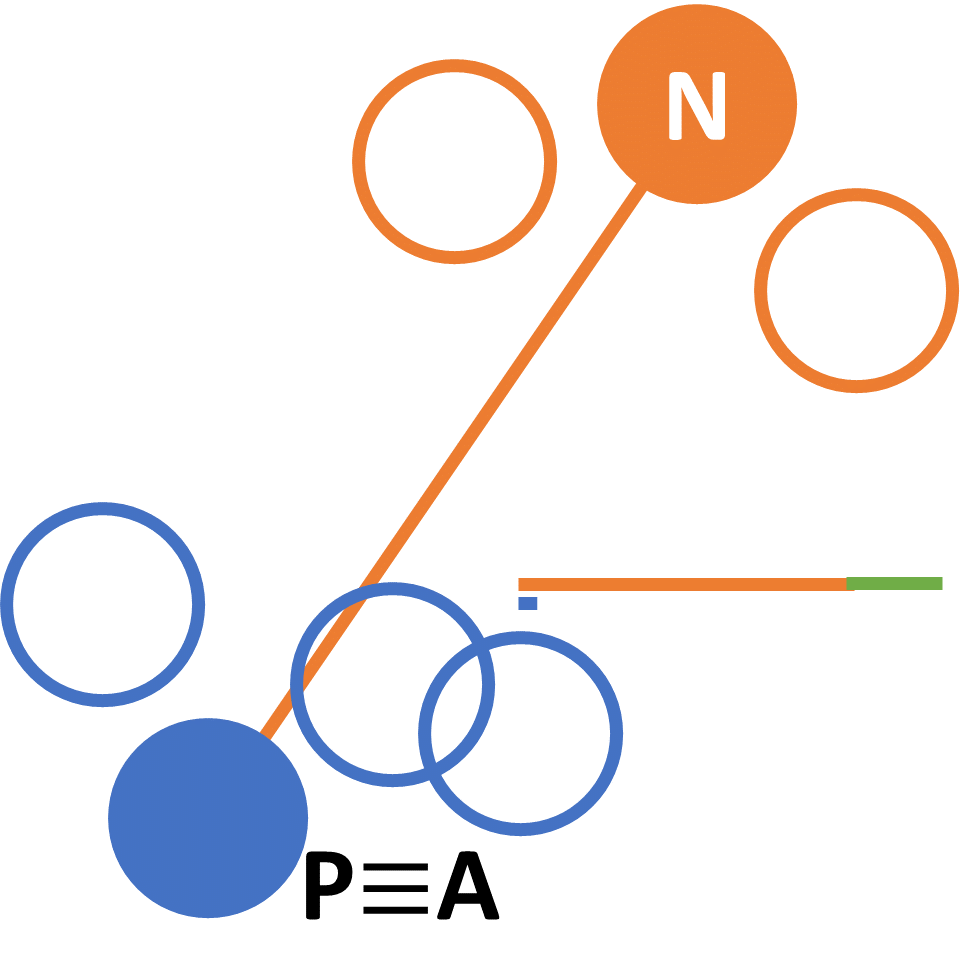}}
}}
\hspace{0.5cm}
\sidesubfloat[]{{
    \frame{\includegraphics[width=0.3\columnwidth]{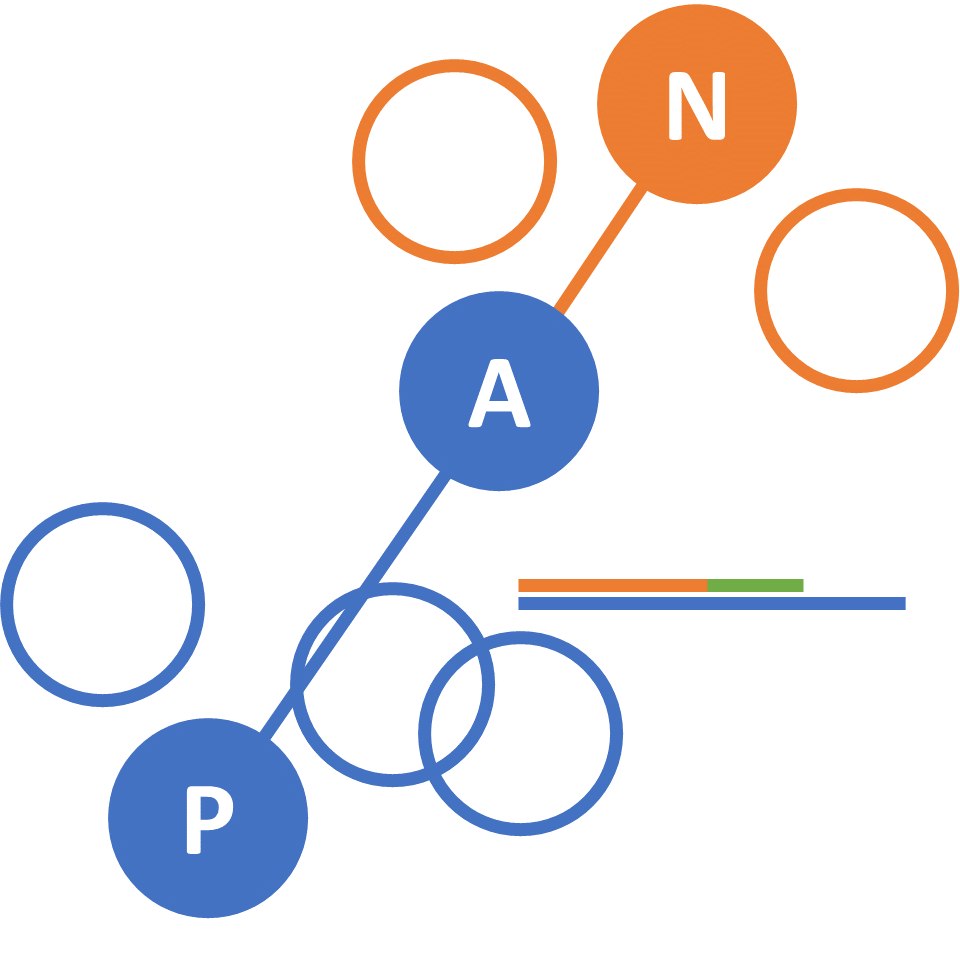}}
}}
\caption{Low dimensional representation of how embeddings are moved across the feature space. In (a), since $d(a,n) + m > d(a,p)$, then $L_{ADV} > 0$. When the anchor is close enough to the negative example, as in (b), the loss becomes $0$.}
\label{img:APNgraph}
\end{figure}

Given the formed triplet, the cloaking algorithm iterates until either the loss function returns $0$ or the anchor is moving towards the negative example too slowly. Fig. \ref{img:APNgraph} shows a lower-dimensional example of this process, in which the negative example $N$ is chosen as the farthest point from $A$ (initially $A \equiv P$). A loss greater than $0$ indicates that the anchor is not closer to the negative example by at least the specified margin than it is to the positive example, so the face can still be further cloaked.

\noindent\textbf{Performance}\\
Using only a single triplet comes with some performance advantages. The main one is the initial choice of examples: looking for a single farthest sample, executed on a processor that supports vector operations, is a fast operation. On a low-powered laptop, it takes on average $2s$ to cloak a face image including pre-processing times, which only take a fraction of a second (see \cite{mtcnn_speed}).

\begin{figure}
\includegraphics [width=0.8\columnwidth]{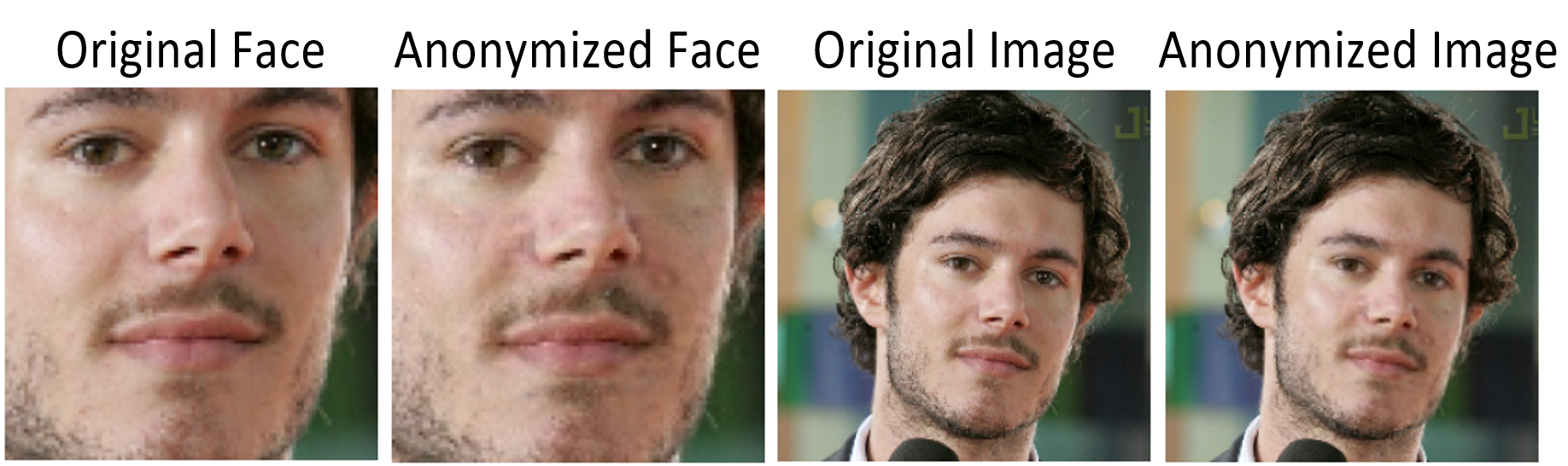}
\caption{\label{img:anonymizedImage}
    Effects of cloaking  as described in method 1.}
\end{figure}

We ran the cloaking algorithm on each of the $250$ samples in our dataset. On average, noise masks show a cumulative alteration of about $177$ pixel values per image or $1.2 \times 10^{-2}$ units per color per pixel (on a $[-1,1]$ range). Such alteration is minimal, as a difference of the equivalent of less than $2$ units per channel on a $[0,255]$ scale for each pixel is almost unnoticeable by the human eye. Fig. \ref{img:anonymizedImage} shows how a face looks before and after being cloaked.

In terms of \textit{Structural Dissimilarity} (DSSIM), the average value registered across all margins is $0.01$. This is significantly lower than what is noticeable with a naked eye ($0.2$), so on average, the alteration is minimal and unnoticeable. The maximum DSSIM registered is also well below the noticeable threshold, sitting at $0.035$.

\noindent\textbf{Efficacy}\\
Results show that the \textit{True Positive} rate is around $0.4$ for all margins used, so less than half of the samples are correctly verified. For comparison, the \textit{True Positive} rate for the original, uncloaked $250$ samples is $0.91$. Cloaked images are therefore over $50\%$ less likely to be correctly verified.

Opposite to the reduction in \textit{True Positive} rate is an increase of the \textit{False Positive} rate. The original, uncloaked images yield a $0.06$ rate, which means $6\%$ of the mismatching pairs are wrongly verified, and this number increases to $0.08 \sim 0.11$ when images are cloaked. In relative terms, $33\%$ to $83\%$ more pairs of faces are mistakenly labeled.

\begin{figure}%
\centering
\sidesubfloat[
]{{
    \includegraphics[width=0.3\columnwidth]{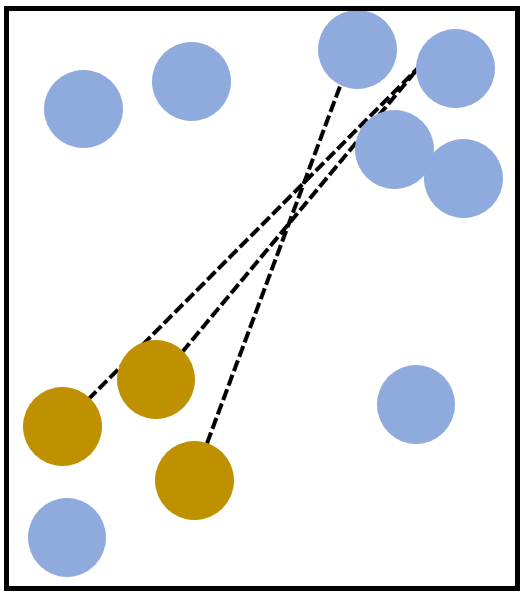}
}}%
\qquad
\sidesubfloat[
]{{
    \includegraphics[width=0.3\columnwidth]{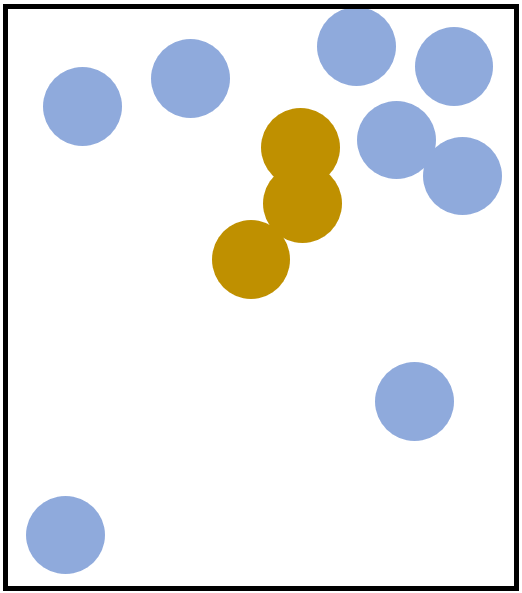}
}}%
\caption{Example of a failed cloaking attempt when selecting the most dissimilar data samples as the target: the embeddings of images of the same person (in yellow) have actually moved closer from (a) to (b), even if each in (b) is supposed to be anonymized.}%
\label{img:FailedClustering}%
\end{figure}

What is surprising is that as the margin grows larger the \textit{True Positive} rate increases. This is probably due to the embeddings being squeezed in the same region of the output hyperspace. An example of this in a lower-dimensional space is given in Fig. \ref{img:FailedClustering}. This is further validated by the \textit{False Positive} rates, which increase as well.

\subsection{Method 2 - Random Samples}


As validated by the efficacy scores, the previous cloaking strategy tends to squeeze some embeddings onto the same area(s) of the model's learned manifold, which could lead to unsupervised classification methods being still able to form clusters of people's faces.
A possible solution is to pick negative examples randomly. The main advantage over selecting the farthest sample is that the embeddings of faces of the same person are likely to be moved in different directions, thus making it more difficult to cluster them. A drawback is that, depending on which sample is picked, the noise mask applied on the face may not be particularly effective.

\noindent\textbf{Algorithm}\\
To mitigate poor choices of the negative example we select multiple examples for a single anchor, thus forming multiple triplets. The cloaking algorithm is executed as in the previous section but once for every triplet. We use $k=5$ for the number of random negative examples to select from the dataset, resulting in 5 triplets.

The noise masks applied on the faces are cumulative across cloaking iterations with each triplet. Once a face is fully cloaked relative to a triplet, the resulting cloaked face is used in the next triplet. The positive example, or the original face's embedding, is kept fixed across all triplets: this ensures that no matter what negative example is chosen, the anchor will always move farther away from its original position, and never closer.

\noindent\textbf{Performance}\\
Cloaking faces using multiple negative examples comes at a slightly higher computational cost. Because cloaks applied on face images are cumulative, the effects of each partially carry over across triplets, and it becomes increasingly faster for the algorithm to converge for a triplet. The computational cost in terms of time required to process each face image is increased by around $50\%$, from $2s$ to $3s$ on the same low-spec machine. The DSSIM scores also see an increase, sitting on average at $0.016$, but still far below the $0.2$ visible threshold.

\noindent\textbf{Efficacy}\\
The verification accuracy in terms of \textit{True Positive} rate is significantly lower than that of the previous method, with values ranging from $0.37$ down to $0.21$, for a relative decrease from $59\%$ to $77\%$. This shows that selecting multiple random targets is more effective than selecting a single farthest one.
The \textit{False Positive}, on the other hand, oscillates between $0.07$ and $0.09$, for a relative efficacy of up to $50\%$ (as up to $3\%$ more samples are mistakenly considered matching).

\subsection{Method 3 - Random \& Farthest Samples}

When $5$ random negative examples are chosen, the probability that they all belong to the same person as the anchor is as low as $10^{-9}$ using our validation dataset, but they are nonetheless not guaranteed to be much dissimilar from the anchor. To mitigate the drawbacks of the previous methods we propose a union of the two: selecting $k$ random negative examples and adding to them the most dissimilar one.

\noindent\textbf{Algorithm}\\
The algorithm is almost identical to the previous ones, the only difference being that an additional, sixth triplet is used, containing the farthest sample. This triplet is processed first, then the ones containing random examples follow suit. The main intuition behind this combined strategy is to first cloak a face with maximum efficacy, and then add additional random (but still targeted) noise to reduce the chances of clustering accurately, as depicted in Fig. \ref{img:FailedClustering}.

\noindent\textbf{Performance}\\
The time required to process each image is on average similar to that of the previous algorithm, which was using $5$ instead of $6$ triplets. The DSSIM scores also remain similar, sitting at $0.016$ on average.

\noindent\textbf{Efficacy}\\
Surprisingly, this strategy yields worse changes in the \textit{True Positive} rate than any previous algorithm, with values from $0.37$ and $0.52$: the best efficacy for this strategy is no better than the worst efficacy of any previous one.
The \textit{False Positive} rate ranges from $0.08$ and $0.095$, with an average of $0.085$. This is not significantly better than any previous strategy. Considering both the performance and the efficacy, this algorithm is the worst of so far.

\subsection{Method 4 - Most Similar Sample}

Performing experiments using a strategy that seemed logically better than others actually yielded the worst results. The following method on the other hand, while counter-intuitive, yields better results.

\noindent\textbf{Algorithm}\\
Instead of using the most dissimilar or farthest away embeddings as target(s), this algorithm uses the most similar examples from the dataset. As explained in section \ref{par:ExampleSelectionAndMargin}, in order for $L_{ADV}$ to be differentiable (and therefore for $\nabla_X L_{ADV}$ to be computable), $d(a,n) < d(a,p) + m$ must be satisfied, where $d$ is the Euclidean distance between two faces' embeddings. As $p \equiv a$ in the first iteration, $d(p,n) > m$ becomes the condition for $\nabla_X L_{ADV}$ to point in a meaningful direction. $n$, or $\hat{x}$, is therefore chosen as the most similar sample by picking the embedding closest to the anchor, such that it is distant from the anchor by at least the margin $m$:
$$\hat{x} = \argmin_{\hat{x} \in D} \left\{\norm{f(x)-f(\hat{x})}_2 \mid \norm{f(x)-f(\hat{x})}_2 > m\right\}$$

\noindent\textbf{Performance}\\
Using the most similar (and yet far enough) sample from the dataset yields similar or better performances compared to previous strategies. As far as computational time goes, the time required to process images decreases slightly to an average of $1.7s$ per sample, a $15\%$ decrease from the first algorithm.

What shows a more significant improvement, on the other hand, is the DSSIM achieved in this algorithm. It is $7.4 \times 10^{-3}$ on average, a $29\%$ reduction, when compared with the best average score, achieved so far. The maximum value registered is below $0.035$, which is acceptable by a large margin. These scores are not particularly surprising, as cloaking a face image using a similar image as the target would logically produce less visible noise masks.

\noindent\textbf{Efficacy}\\
What is instead an unexpected result is the efficacy of this algorithm. The \textit{True Positive} rates vary greatly depending on the margin used, from $0.87$ at margin $0.2$ (which is an insignificant accuracy reduction) down to $0.08$ as the margin approaches $2.0$. As the margin increases, in fact, the \textit{True Positive} rate decreases to results that not even a random target selection could achieve. In relative terms, the accuracy drops by up to $91\%$ compared to the verification of pairs of original images. This is by far the best result. 
The \textit{False Positive} rate for this strategy is also largely dependent on the value of the margin and oscillates from a minimum of $0.07$ with $m=0.2$ and a maximum of $0.11$ with $m=1.4$.

\subsection{Method 5 - No Sample Selection}
\label{par:no_sample_selection}
The best algorithm analyzed so far is also the one expected to perform the worst. We, therefore, push the idea of selecting a similar sample as a negative example to the extreme by picking the anchor itself. While at first glance it may seem counter-intuitive to cloak an image using as target the image itself, we provide analytical reasoning to support its efficacy.

When a negative example, or target, is used, the gradient of the loss function indicates how to alter the input to make it more similar to the target, or how to minimize the loss in relation to the target; on the other hand, if a target is not used, the gradient is used to maximize the loss as calculated with the true class. The latter case can be translated in the context of feature extractors as maximizing the representational inaccuracy of a data sample, or a facial embedding in our case.

\noindent\textbf{Algorithm}\\
Simply enough, this algorithm requires the anchor to be selected as both the positive and the negative example at the first iteration. Additionally, an insignificant $\varepsilon$ is added to the negative sample to make it just slightly different from the positive one and maintain differentiability for $L_{ADV}$. The $\varepsilon$ difference added to $n$ is a small increase of $1e^{-5}$ to the pixel values of an image $x \in [-1,1]^{160\times160\times3}$. This does not actually have any effect on the images, because if it were converted into the standard $[0,255]$ range, no pixel would be altered. 

Upon carefully inspecting $L_{ADV}$, we have found that setting $p \simeq n$ produces a gradient $\nabla_x L_{ADV}$ that points $a$ in a seemingly random direction, but always away from the original samples $p$ and $n$. This is because, following Eq. \eqref{eq:l_adv} and using $d$ as defined in Eq. \eqref{eq:phi}, the adversarial loss becomes
\begin{equation}
\begin{split}
L_{ADV} (x+\epsilon, x, x+\varepsilon) = \\
d(x+\epsilon, x) - d(x, x+\varepsilon) + m
\end{split}
\label{eq:no_target_loss}
\end{equation}
and its gradient w.r.t. the input is calculated as a small semi-random vector that introduces instability in the network. $\nabla_x L_{ADV}$, in this case, becomes an alteration that pushes the anchor away from its origin but not towards any target. As the model's manifold most likely contains regions that do not correspond to any existing face image and given its high dimensionality, $a$ is likely pushed towards regions of it that would be otherwise unreachable following previous strategies, because no face image maps to there. We, therefore, exploit the intangibility of the model's manifold to produce adversarial examples that other methods do not yield.

\noindent\textbf{Performance}\\
This version of the algorithm takes more time than some of its previous iterations, with an average time to process an image of $3s$. The intensity of the alterations, on the other hand, is the lowest so far. The average DSSIM score is $0.01$, about $4\%$ lower than that using the farthest sample as $n$, with a maximum of $0.03$ - also slightly lower than that of the first algorithm.

\noindent\textbf{Efficacy}\\
The most interesting result of this algorithm is the \textit{True Positive} rate when verifying pairs of images with matching identities: $0.084$, which means only $8.4\%$ of the pairs are correctly verified. This is a hindering of $90\%$ of the verification accuracy compared to clear images, the best result among all strategies. 
Despite the \textit{True Positive} rate being very low, the \textit{False Positive} rate is low, too, even more than that of uncloaked images: $0.027$. This is explained by how embeddings are scattered in semi-random regions of the manifold and are thus not likely to end up close to one another.

\subsection{Results Summary}

Of the five cloaking strategies analyzed, selecting the same positive and negative examples proved to be the optimal choice. While each strategy's outcomes were described in detail and included informative comparisons, a final summary of the results can help to better understand how our strategies compare to one another.

The first comparative analysis proposed is that of the DSSIM scores registered. For each of the five algorithms and each of the $10$ values proposed for the margin, we have found the average, maximum, and standard deviation of the DSSIM values between pairs of cloaked and uncloaked images. As the margin increases from $0.2$ to $2.0$, the mean DSSIM scores increase linearly from $0.01$ to $0.015$. The standard deviation is in all cases within $0.005$, and the maximum values registered are outliers sitting at $0.05$, found using the \textit{random} and \textit{random \& farthest} strategies with margins $m>1$.

\begin{figure}[t] \includegraphics [width=0.7\columnwidth] {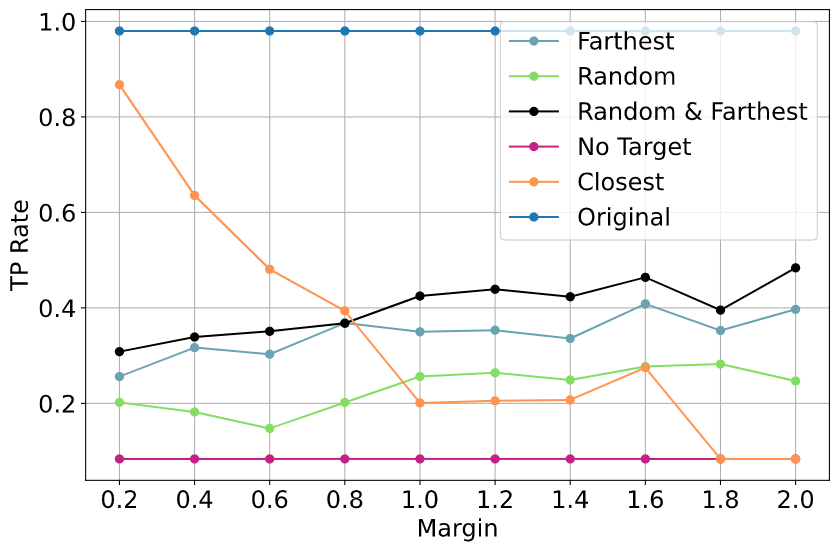}
\caption{\label{img:dev-tp}
    True Positive rates for each experiment (including a baseline computed with uncloaked face images).
}
\end{figure}

An important graph to analyze is that of the \textit{True Positive} rates in Fig. \ref{img:dev-tp}, which includes a data point for each combination of algorithm and margin value. The graph shows a general and slight increase in verification accuracy as the margin increases for algorithms based on cloaking images by selecting examples far away. In contrast, selecting nearby samples as targets yields better results as the margin is set to be larger. The efficacy of the closest samples selector is the same as selecting no targets when the margin is $\geq 1.8$. This is because in most cases, given a face image there is no other example that is at least $1.8$ far from it, and the closest sample is then set to the anchor itself, as in the last strategy proposed.

\begin{figure}[t] \includegraphics [width=0.7\columnwidth] {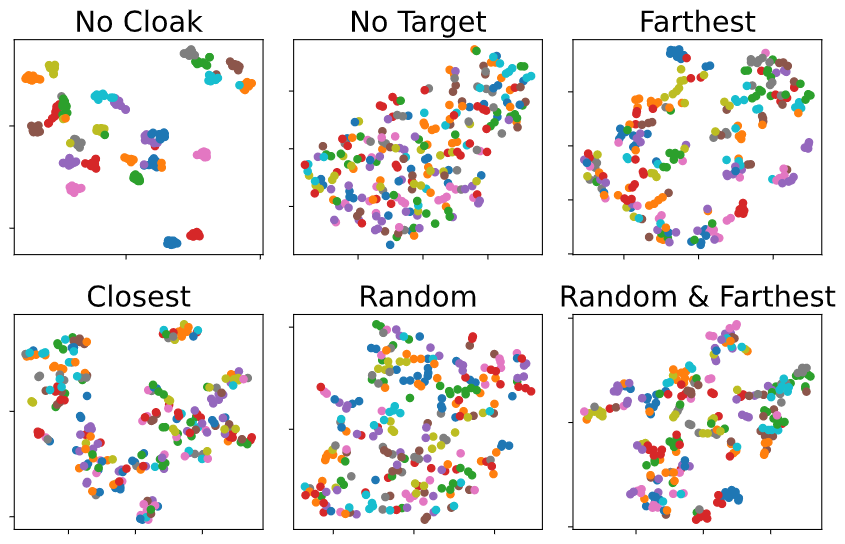}
\caption{\label{img:dev-tsne}
    Visualization on a 2D plane of the embeddings using t-SNE. The embeddings are cloaked using a margin of $1.0$. Different colors correspond to different people's identities.
}
\end{figure}

A final visualization of the effects of cloaking images with different techniques is given by the graphs in Fig. \ref{img:dev-tsne}. Each scatter plot shows the results of performing \textit{T-distributed Stochastic Neighbor Embedding} (t-SNE) \cite{t-SNE} on the embeddings. This is a non-linear technique to reduce high-dimensional vectors to a smaller subset of features. t-SNE is very similar to \textit{Principal Component Analysis} in concept, and it yields the best results when processing data with a large number of features, such as embeddings. The top-left graph shows how embeddings are mapped onto two dimensions, with each point representing an embedding and being colored according to the identity of the person portrayed. Not surprisingly, the original face images' embeddings form very distinct clusters. Once images are cloaked, however, their embeddings are more scattered. It can be seen that the first algorithm proposed produces embeddings that still form clusters, to some extent. Randomly picking $5$ targets as negative examples produce more scattered results, while the best cloaking results are given by setting the negative and positive examples equal to one another: this latter case does not allow any clusters to be inferred.

\vspace{-.5cm}
\section{Evaluation}

This section proposes a complete evaluation of the effects of cloaking images with Ulixes for three different tasks: \textit{verification}, \textit{supervised identification}, and \textit{unsupervised identification} (term hereafter used interchangeably with \textit{clustering}). For each of these tasks, we use an appropriate dataset. We test verification on the Labeled Faced in the Wild (\textit{LFW}) \cite{LFW} dataset, and test identification and clustering on the \textit{FaceScrub} \cite{FaceScrub} and \textit{PubFig83} \cite{PubFig83} datasets. An additional subsection then compares the generalization of Ulixes to other models and compares its transferability with Fawkes \cite{Fawkes} and Face-Off \cite{face-off}.

\label{par:accuracy}
\textbf{Verification} is the task of determining if two images represent the same person or not. The LFW \cite{LFW} dataset provides $3000$ pairs of images of the same people and $3000$ of different people. Matching pairs can be verified successfully (True Positive) or not (False Negative), and so pairs of mismatching identities can be successfully rejected (True Negative) or wrongly verified (False Positive). We hereby refer to these terms as $TP, FN, TN$, and $FP$. Accuracy is here defined as $(TP + TN) / (TP + FN + TN + FP)$. A real-world application of verification is checking an employee's identity given his ID card before entering a corporate building.

\textbf{Supervised Identification}, hereafter also simply called \textit{identification}, is the task of, given a labeled dataset of faces, finding the label of new face images whose real identity is among those contained in the dataset. A practical implementation of this is Adobe Lightroom's facial recognition tools \cite{lightroom}. Accuracy is here calculated as the ratio of faces correctly classified.

\textbf{Unsupervised Identification}, or \textit{clustering}, is the task of trying to label a fairly large number of images with no prior knowledge of them. An actual application of this would be to separate a set of images in groups containing each the same person when no labeled images for those people are available.

Note how each subsequent task uses less information than the previous. Verification is a true or false problem which, given a large amount of data (two images), tries to predict only one bit of information. Identification, on the other hand, is a classification task, and its prediction is more complex than just a yes or no: the more classes are available, the harder the problem becomes. This said, identification is advantaged by the usage of a labeled dataset, and therefore it can take advantage of  prior information. Finally, clustering is similar to identification in terms of prediction goals; however, it does not have a labeled dataset for reference. These conditions make it the task that offers the least amount of prior knowledge.

We, therefore, assess the efficacy of our best performing algorithm for each of the levels of information that an entity performing facial recognition may have available: a single image for each person, on which only verification is to be done; multiple labeled images of many people; multiple, unlabeled images of many people.

\subsection{Verification}

We use the LFW dataset with its lists of matching and mismatching pairs of images to assess the effects of the cloaks applied to images. Verification is done by comparing two images' embeddings by computing their Euclidean distance and classifying them as a match if their distance is less than a threshold. Different thresholds yield different results: a low threshold would yield a very low $FP$ rate, but also a low $TP$ rate; setting it to a higher value should dramatically increase the $TP$ rate, but at the cost of registering higher $FP$ rates too.

\begin{figure}
\centering
\includegraphics [width=0.65\columnwidth] {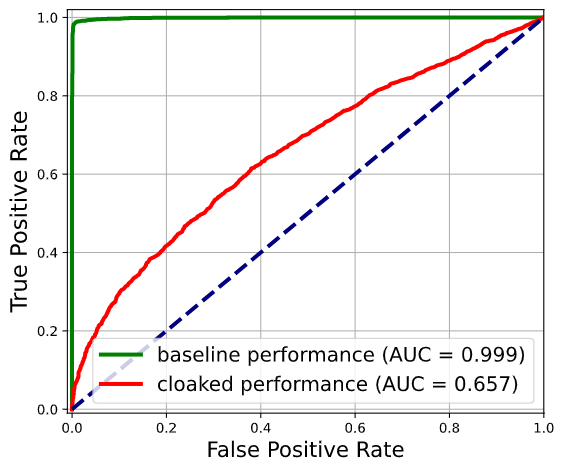}
\caption{\label{img:roc}
    Comparison of the verification efficacy on LFW between the original and the cloaked images. The accuracy with uncloaked images is $99.01\%$, whereas cloaked images give $61.75\%$.
}
\end{figure}

Fig. \ref{img:roc} shows the \textit{ROC} curve typically used to evaluate the efficacy of a facial feature extractor for the verification task. The graph is obtained by setting the threshold to $0$ and increasing it with steps of $0.001$ until it reaches the value of $3.0$ (for a total of $150$ values) and calculating the true and false-positive rates for the pairs of embeddings used. For comparison, two curves are plotted: the first is calculated using the embeddings as extracted from the original images, while the second is calculated on those of the cloaked images. While verification is not perfect on the original face images, it yields quite a high accuracy while keeping the \textit{False Positive} rate to a minimum. When images are cloaked, however, verifying whether two images represent the same person or not is almost as accurate as a coin toss. The cloaks applied to the images have, therefore, successfully made it difficult to extract information from them via verification alone.

\subsection{Supervised Identification}

This experiment is used to evaluate the efficacy of Ulixes against various identification methods. We explore five different identification strategies, so to make sure results are assessed as thoroughly as possible. The following algorithms are used for supervised identification:
\begin{itemize}
    \item[-] Nearest Neighbor (nn)
    \item[-] Nearest Centroid (nc)
    \item[-] K-Nearest Neighbors (knn)
    \item[-] Weighted K-Nearest Neighbors (wknn)
    \item[-] Distance-Bound Nearest Neighbor (dbnn)
    \item[-] A fully connected neural network with two hidden layers, each having 20 neurons, which is trained until its training accuracy is above $95\%$ (FC net)
    \item[-] A multi-class support vector machine (SVM)
\end{itemize}
\vspace*{-\baselineskip}
\vspace{0.2cm}

Before being able to effectively evaluate the effects of cloaking images in identification tasks, it is necessary to define a comprehensive experimental setup. We have analyzed the accuracy of the previously listed identification strategies with respect to the breadth and depth of a labeled dataset. For the breadth, which in this case corresponds to the number of people included in the labeled dataset, we have chosen values ranging from $2$ to $30$. For depth, which is how many samples are labeled for each person, we have chosen values $2$, $5$ and $15$.
\begin{figure}
\includegraphics [width=0.7\columnwidth] {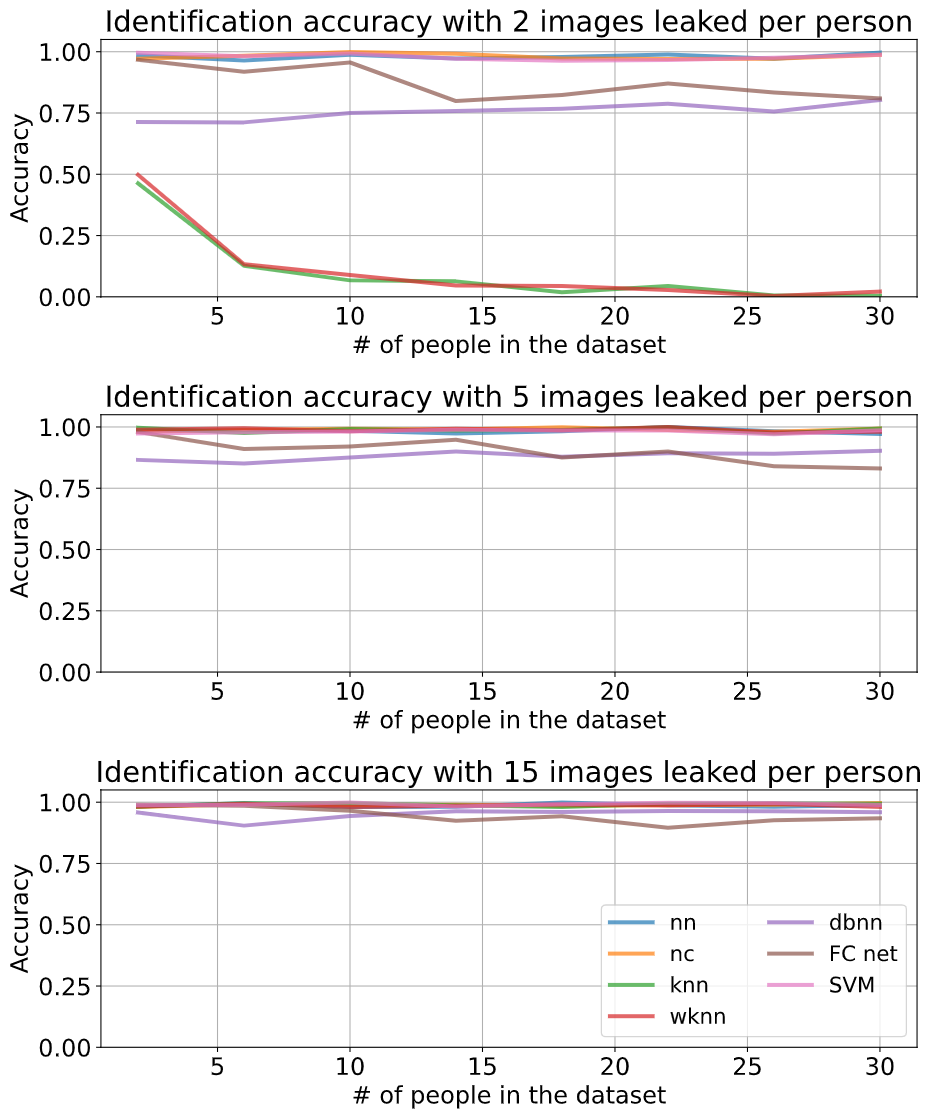}
\caption{\label{img:id_test}
    Supervised classification accuracy of 7 identification methods varying the depth and breadth of the labeled dataset used for training/reference. Note that the values in the ordinate are in the range $[0.6,1]$ to increase resolution.
}
\end{figure}

The graphs in Fig. \ref{img:id_test} show how the accuracy of identification, calculated as the number of correctly identified images over the number of images processed, for uncloaked images. Each of the three graphs shows a different value for the depth of the labeled dataset, while the breadth of the said dataset is varied on the abscissas. The accuracy of the fully connected network and the distance-bound nearest neighbor take a hit when the number of available training samples (the depth of the labeled dataset) is low. However, as the dataset grows deeper, the accuracy increases for all analyzed strategies. These results are used as a baseline to compare how cloaking images affects supervised classification.

\begin{figure}
\includegraphics [width=0.7\columnwidth] {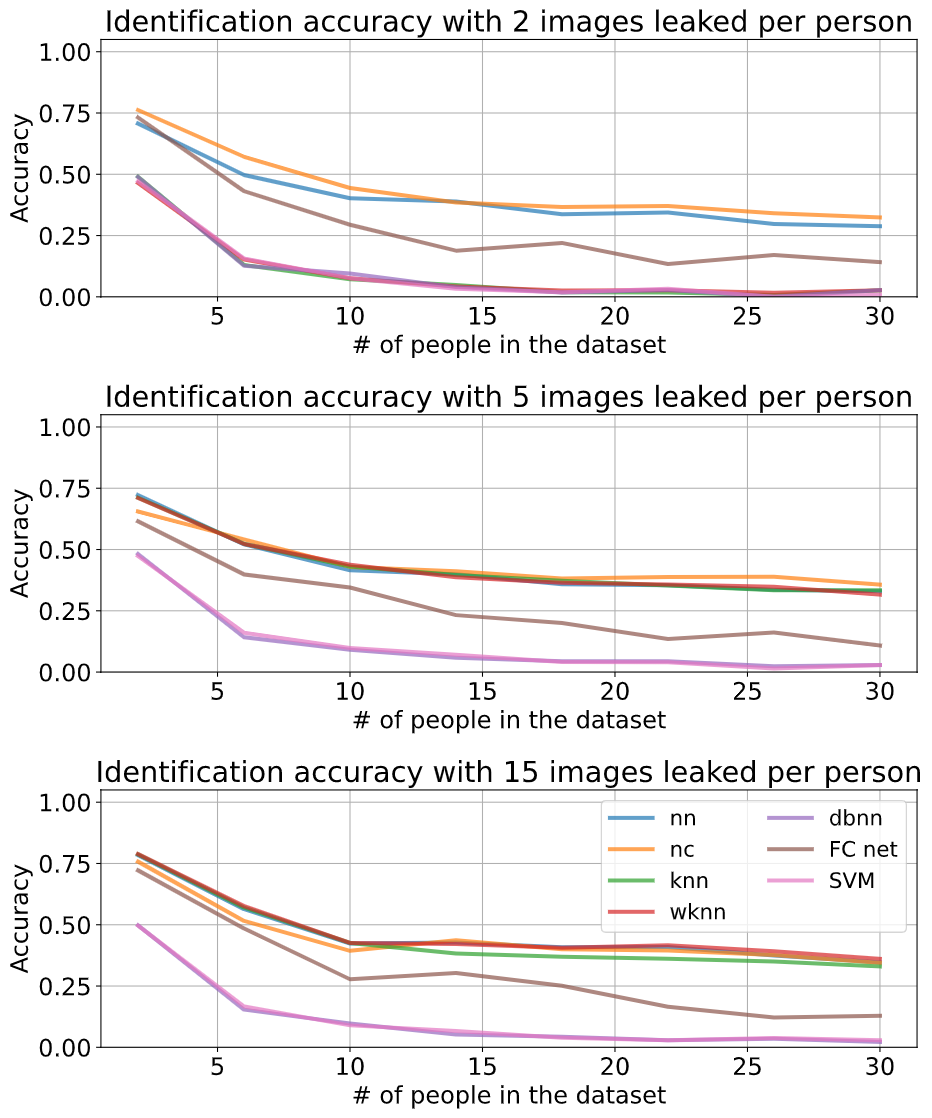}
\caption{\label{img:id_test_cloaked}
    Repetition of the experiments in Fig. \ref{img:id_test} using cloaked images as samples to classify.
}
\end{figure}

Once images are cloaked, supervised identification strategies fail to properly classify faces. Fig. \ref{img:id_test_cloaked} shows the results of the same experiments conducted with cloaked face images as samples to be classified. In this setup, the same uncloaked face images are used to train the identifiers, and cloaked images are then classified. This model is representative of an attacker with the following capacities: the training dataset of the encoder (FaceNet) is trained on uncloaked images, including some of the people that are later classified; the attacker has $2$ to $15$ uncloaked sample images per person to be identified; the images to be classified are cloaked. The results show that the accuracy drops considerably for almost all breaths and depths of the labeled dataset. When the breath of the dataset is $2$, the accuracy remains fairly high ($\approx 75\%$), but as soon as the breath increases, the accuracy drops. As separating samples across $2$ classes is a task that is unlikely of any practical use to an attacker, the lower efficacy of the cloaks in this challenging setup is negligible.

\subsection{Clustering}\label{sec:clustering}
The final evaluation for the effectiveness of Ulixes is done via clustering, one of the most commonly used unsupervised learning techniques. The goal is to look at whether it is possible to group together face images of the same individuals. As previously done, we first look at how well clustering works when performed on the original, unaltered face images. We then compare this to when $10\%$ of the images, picked at random, are cloaked, then when $50\%$ are, and finally when only cloaked images are used. Measurements are also results of a tenfold average for datasets with $30$ people having each $30$ images.

The clustering accuracy of a method is calculated by looking at the individual clusters. For each cluster, its most frequent item is taken as the cluster identifier; then, the percentage of images in the cluster belonging to the same person is calculated, and a weighted average is taken for identified clusters (excluding clusters with a single item), where the weight is directly proportional to the cluster size. This measurement allows a single person's images to be split into two clusters but still yield a $100\%$ accuracy if both clusters only include images of that person and are not outliers, thus allowing large differences in a person's look across images (e.g., haircuts and aging effects) to not negatively affect the score.

This evaluation is performed using agglomerative clustering with one of two stop conditions: either when subsequent agglomerations exceed a maximum distance threshold of $1.242$ (which allows the formation of multiple clusters per person), or when the target number of clusters is reached ($30$, which means that the clustering algorithm is given some information about the composition of the dataset). As a consequence, Table \ref{tab:clustering} reports two measurements for each experiment, one per stop condition of the clustering algorithm. The resulting accuracy scores show a significant decrease as the percentage of cloaked images increases. When all images are cloaked, compared with when they are not, faces are over $70\%$ less likely to be correctly clustered, which is a significant result.
\begin{table}[t]
\centering
\caption{Clustering accuracy scores on two datasets, PubFig83 and FaceScrub, as different ratios of images are cloaked. The first configuration 'Config 1' refers to clustering with $d=1.242$ stop condition, the second with a stop condition dependent on the number of clusters formed ($30$). The accuracy reported is calculated as explained in the second paragraph of section \ref{sec:clustering}, with the values in parenthesis being the relative decrease from the baseline (when $0\%$ of the samples are cloaked).}
\label{tab:clustering}
\resizebox{0.85\columnwidth}{!}{%
\begin{tabular}{c|c|c|c|c} 
& \multicolumn{2}{c|}{PubFig83} & \multicolumn{2}{c|}{FaceScrub}  \\ \cline{2-5}
& Config 1  & Config 2        & Config 1    & \multicolumn{1}{c|}{Config 2}          \\ \hline
\multicolumn{1}{|c|}{0\%}     & 98.41\%     & 98.60\%         & 99.50\%     & \multicolumn{1}{c|}{99.86\%}           \\
\multicolumn{1}{|c|}{cloaked} &             &                 &             & \multicolumn{1}{c|}{}                  \\ \hline
\multicolumn{1}{|c|}{10\%}    & 88.46\%     & 85.85\%         & 90.84\%     & \multicolumn{1}{c|}{87.16\%}           \\
\multicolumn{1}{|c|}{cloaked} & (-11.01\%)   & (-12.93\%)      & (-8.70\%)   & \multicolumn{1}{c|}{(-12.64\%)}        \\ \hline
\multicolumn{1}{|c|}{50\%}    & 56.82\%     & 48.77\%         & 62.06\%     & \multicolumn{1}{c|}{51.23\%}           \\
\multicolumn{1}{|c|}{cloaked} & (-42.84\%)  & (-50.54\%)      & (-37.63\%)~ & \multicolumn{1}{c|}{(-48.69\%)}        \\ \hline
\multicolumn{1}{|c|}{100\%}   & 26.14\%     & 23.09\%         & 28.98\%     & \multicolumn{1}{c|}{26.20\%}           \\
\multicolumn{1}{|c|}{cloaked} & (-73.70\%)  & (-76.58\%)      & (-70.87\%)  & \multicolumn{1}{c|}{(-73.68\%)}        \\
\end{tabular}
}
\end{table}

\section{Generalization \& Transferability}
In this section, we consider two essential aspects of an adversarial algorithm: generalization and transferability.  Generalization is the ability of an algorithm to be used with different underlying machine learning models and training sets. An adversarial perturbation to a data sample is said to be transferable when the same perturbation is effective against different models. This is of crucial importance in the design and evaluation of an adversarial machine learning algorithm. White-box attacks, while often more powerful than black-box attacks, have the drawback that if the attack is not transferable, then it is only effective as long as an attacker has access to its victim. In our threat model, where online services' users are the victims and facial recognition practitioners are the attackers, we assume the attackers use models and training datasets unknown to the users. It is, therefore, necessary for users to use cloaks that are effective not only against a single facial recognition system but multiple.

In order to measure the generalization and transferability of Ulixes, we conduct a series of experiments. The first set of experiments measures the performance of our algorithm with different models and training datasets, and the transferability of cloaks generated with a particular model trained on a specific uncloaked dataset to other models, possibly trained on different datasets. This gives a tangible measure of whether the manifold learned by different models is similar in those regions that contain face images, and whether the effects on one propagate onto another. Next, inspired by Fawkes' experimental setup \cite{Fawkes}, we test our cloaks against the online API Face$^{++}$ \cite{face++}. 

All tests within this section are also performed using images cloaked using Fawkes\footnote{https://github.com/Shawn-Shan/fawkes} and Face-Off. \footnote{https://github.com/wi-pi/face-off} Fawkes operates in its second-most invasive mode, \textit{mid}, which sets its maximum number of steps to 75 and the threshold for distance to 0.012 (see its source code for further explanations). On the other hand, Face-Off is run with its default parameters, such as amplification set to 5.1 and kappa 6.0, and a batch size of 30.

\subsection{Cross-Model Transferability}

\cite{transferability} proposes a distinction between \textit{intra-technique} and \textit{cross-technique} transferability, depending on whether adversarial examples are transferred between models of the same kind (kNNs to kNNs, or DNNs to DNNs), or of different nature (e.g. when adversarial examples generated on an SVM are transferred to an LSTM). In the context of this research, due to the high complexity of the models used, it is difficult to define whether two are of the same kind or not. We, therefore, refer to our experiments as studying the \textit{cross-model} transferability of adversarial examples from a facial recognition system to another.

We use four facial recognition models to generate our cloaks, of which one is trained on different datasets, for a total of five configurations. The first two types are trained with the \textit{triplet loss} function, while the latter with the \textit{cross-entropy loss} function. These are:
\begin{itemize}
    \item[-] Facenet \cite{FaceNet}, trained on both the CASIA-WebFace dataset \cite{cwb} \cite{facenet_models} and the VGGFace2 dataset \cite{vggface2};
    \item[-] VGGFace, in its VGG-16 architecture \cite{vgg-face} and ResNet50 architecture \cite{vggface2};
    \item[-] Facebook's Deepface \cite{DeepFace_1}.
\end{itemize} 
\vspace{-0.4cm}

For each configuration, we use Ulixes to cloak all faces in the LFW dataset \cite{LFW}. The transferability of the generated adversarial examples is assessed by measuring the accuracy of each of the five models (as defined in Section \ref{par:accuracy}) in verifying the matching and mismatching pairs proposed by the LFW benchmark. \footnote{http://vis-www.cs.umass.edu/lfw/} Additionally, we use Fawkes and Face-Off to similarly cloak the LFW dataset, and compare the performances of the three systems.

Table \ref{tab:transferability} shows the transferability matrix resulting from the experimental setup described above. Adversarial examples generated on a model transfer well to the same model trained on a different dataset. Such examples also show high transferability across different architectures of VGGFace. However, little to no \textit{cross-model} transferability is registered. We believe this is due to the manifolds that each model has learned: even given partially overlapping training datasets, different systems have modeled a different representational manifold. Additionally, since different models require scaling images to specific sizes, resizing could negatively impact transferability. Both Fawkes and Face-Off perform poorly, being unable to successfully transfer attacks to any model. While these results are not surprising for Fawkes, since it is best suited to poison training datasets, Face-Off introduces clearly visible noise overlays in images and it is interesting how little of an effect they actually have.

\noindent\textbf{Ensemble Methods}\\
In an attempt to improve cross-model transferability, we propose three ensemble methods:
\begin{enumerate}
    \item The first consists of overlaying the noise masks generated by three different models onto the same image. While trivially logical, overlaying cloaks do not simply result in perfect transferability. This is because Ulixes may change a pixel in a direction to hinder one facial recognition model, and in another direction to hinder another model. When the two masks are overlayed, they cancel each other out in that pixel. In our experiments, we have found this method, \textit{Ensemble\textsuperscript{1}}, to produce images with more invasive noise masks, but still with DSSIM scores have always remained far below $0.2$. 

    \item A second approach to promote transferability is to take, for every pixel coordinate and color channel in a face image, the one element with the largest absolute value across the cloaks of the models in the ensemble, at corresponding coordinate and color channel. This is referred to as \textit{Ensemble\textsuperscript{2}}.
    
    \item The third approach to ensemble methods is that proposed by Liu et al. \cite{targeted_untargeted}, which we hereby refer to as \textit{Ensemble\textsuperscript{3}}. This consists of combining the gradients of the models in the ensemble at every step of the attack. While in previous methods masks are first fully computed, and then combined, possibly resulting in masks canceling out, \cite{targeted_untargeted} every step taken is optimal.
\end{enumerate}
\vspace{-.4cm}

In our evaluation of the effectiveness of masks generate from ensembles, we exclude the evaluation model from the ensemble.

The results of applying cloaks generated with these ensemble models are also included in Table \ref{tab:transferability}, and they achieve the best cross-model transferability of all systems, far surpassing both Fawkes and Face-Off against all victim models. \textit{Ensemble\textsuperscript{1}} remains effective in cloaking and generalizes well to all models; however, its visual impact is significant. For a few images, in order to be cloaked their DSSIM scores exceed $0.2$, and the average DSSIM is $0.1$. This results in images being visibly altered. On the other hand, \textit{Ensemble\textsuperscript{2}} still produces imperceptible cloaks and its adversarial examples are fairly anonymized. Finally, \textit{Ensemble\textsuperscript{3}} has similar efficacy to \textit{Ensemble\textsuperscript{2}}, and the average and maximum DSSIM scores registered from it are $0.05$ and $0.11$, respectively.

The ensemble method proposed by Liu et al. \cite{targeted_untargeted} requires building a large neural network model that connects the last layers of multiple smaller models. Due to this, its memory footprint is much greater and, if unable to fit in a small memory, can lead to an execution time that scales non-linearly with the number of models used. Its performance is therefore closely dependent on its implementation and the machine it is run on, whereas the \textit{Ensemble\textsuperscript{1}} and \textit{Ensemble\textsuperscript{2}} methods can be used more easily as they do not need to be run in parallel. Due to these considerations and the experimental results reported, we believe that \textit{Ensemble\textsuperscript{2}} is the best method of those discussed.

\begin{table*}
\centering
\caption{Transferability matrix of cloaks from a model/system onto another. The systems in normal font are used to generate the cloaks (attackers), whereas the ones in italic are used to assess such cloaks (victims). The percentages refer to the verification accuracy decrease that adversarial cloaks induce to a victim's model. \textit{FaceNet C-WB} refers to FaceNet trained on the CASIA-WebFace dataset. For the ensemble methods, we exclude the model being evaluated against the ensemble.}
\label{tab:transferability}
\newcolumntype{S}{ >{\centering\arraybackslash} m{0.75cm} }
\newcolumntype{A}{ >{\centering\arraybackslash} m{0.9cm} }
\newcolumntype{M}{ >{\centering\arraybackslash} m{1.2cm} }
\newcolumntype{L}{ >{\centering\arraybackslash} m{1.4cm} }
\resizebox{0.75\columnwidth}{!}{
\begin{tabular}{L|L|M|M|M|S|A|S|M|M|M}
& \textbf{FaceNet VGGFace2 } & \textbf{FaceNet C-WB} & \textbf{VGGFace Resnet50} & \textbf{VGGFace VGG-16} & \textbf{Deep Face} & \textbf{Fawkes} & \textbf{Face-Off} & \textbf{Ensemble\textsuperscript{1}} & \textbf{Ensemble\textsuperscript{2}} & \textbf{Ensemble\textsuperscript{3}} \\ \hline
\textit{FaceNet VGGFace2 }      & -76\% & -76\% & -7\%  & -7\%  & -4\%  & -2\% & -6\% & -55\% & -44\% & -45\% \\ \hline
\textit{FaceNet C-WB }          & -76\% & -76\% & -7\%  & -7\%  & -4\%  & -2\% & -6\% & -55\% & -44\% & -45\% \\ \hline
\textit{VGGFace Resnet50 }      & -6\%  & -6\%  & -27\% & -29\% & -2\%  & -3\% & -5\% & -29\% & -22\% & -26\% \\ \hline
\textit{VGGFace VGG-16 }        & -6\%  & -6\%  & -27\% & -29\% & -2\%  & -3\% & -5\% & -29\% & -22\% & -26\% \\ \hline
\textit{DeepFace }              & -1\%  & -1\%  & 0\%   & -2\%  & -32\% & -3\% & -3\% & -30\% & -28\% & -28\% \\
\end{tabular}
}
\end{table*}


\begin{figure}
\centering
\footnotesize

\stackunder[0pt]{}{
    \begin{minipage}[t]{0.12\textwidth}
        \raisebox{0.7cm}{Original}
    \end{minipage}
    \includegraphics[width=0.18\textwidth]{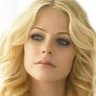}
    \includegraphics[width=0.18\textwidth]{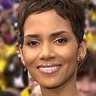}
    \includegraphics[width=0.18\textwidth]{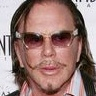}
    \includegraphics[width=0.18\textwidth]{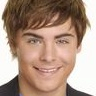}
}%
\\
\stackunder[0pt]{}{
    \begin{minipage}[t]{0.12\textwidth}
        \raisebox{0.7cm}{Ulixes}
    \end{minipage}
    \includegraphics[width=0.18\textwidth]{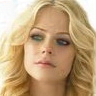}
    \includegraphics[width=0.18\textwidth]{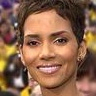}
    \includegraphics[width=0.18\textwidth]{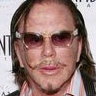}
    \includegraphics[width=0.18\textwidth]{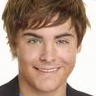}
}%
\\
\stackunder[0pt]{}{
    \begin{minipage}[t]{0.12\textwidth}
        \raisebox{0.7cm}{Fawkes}
    \end{minipage}
    \includegraphics[width=0.18\textwidth]{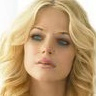}
    \includegraphics[width=0.18\textwidth]{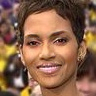}
    \includegraphics[width=0.18\textwidth]{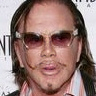}
    \includegraphics[width=0.18\textwidth]{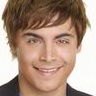}
}%
\\
\stackunder[0pt]{}{
    \begin{minipage}[t]{0.12\textwidth}
        \raisebox{0.7cm}{Face-Off}
    \end{minipage}
    \includegraphics[width=0.18\textwidth]{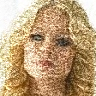}
    \includegraphics[width=0.18\textwidth]{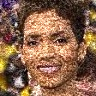}
    \includegraphics[width=0.18\textwidth]{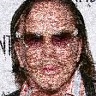}
    \includegraphics[width=0.18\textwidth]{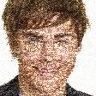}
}
\caption{Some examples of the cloaks generated by Ulixes compared with the same samples cloaked with Fawkes operating in \textit{mid} mode and Face-Off with the default of $30$ training epochs.}
\label{img:examples}
\end{figure}

\subsection{Transferability to Internet APIs}

Inspired by Fawkes' experimental setup, as described in \cite{Fawkes} and on its official project page, we  include an evaluation of Ulixes' performance using Face$^{++}$ \cite{face++}. Face$^{++}$ is an online API that offers facial recognition tools, such as face localization, verification, and identification. Because this is a commercial API, details of the specific implementation are not specified.

Because the threat model proposed in this paper is more challenging than that of Fawkes, the experimental setup is different. We have designed an experiment to assess how cloaks generalize to Face$^{++}$ without poisoning datasets, and we test Ulixes', Fawkes' and Face-off's in the same fashion. For Ulixes' we use the Ensemble MAX algorithm with FaceNet VGGFace2, VGGFace Resnet50, and DeepFace included in the ensemble. The experimental setup is structured as follows:
\begin{itemize}
    \item[-] A labeled dataset with 11,000 images from 3000 classes is uploaded to Face$^{++}$. This is less than a third of the number of classes used in Fawkes, \footnote{https://github.com/Shawn-Shan/fawkes/issues/67} thus making cloaking faces an even more challenging task because there are fewer other samples that a face can be mistaken as;
    \item[-] $3000$ cloaked test images are identified by Face$^{++}$ as one of the available labels, and the predictions are compared with the true classes;
    \item[-] Classification accuracy is calculated as the portion of test samples correctly identified; the cloaks of a system are effective when they reduce the accuracy of Face$^{++}$.
\end{itemize}

Table \ref{tab:face++} summarizes the classification accuracy obtained from images cloaked by each tested system. Fawkes has the lowest performance, as very few of the cloaked images are actually anonymized, which is  partially expected since Fawkes was designed as a data poisoning attack. Face-Off performs roughly four times better than Fawkes, but at the cost of introducing very visible noise masks on face images (see Fig. \ref{img:examples}). Ulixes significantly outperforms both Fawkes and Face-Off and is able to make unrecognizable more than half of the test images.

\begin{table}
\caption{Effectiveness of cloaks against Face$^{++}$.}
\label{tab:face++}
\centering
\resizebox{0.9\columnwidth}{!}{
\begin{tabular}{l|c|c|c|c} 
Cloak System & Baseline & Ulixes & Fawkes & Face-Off \\ 
\hline
Face$^{++}$ Accuracy & 96\% &  45\% &  90\% &  65\% \\
                     &      & (-53\%) & (-7\%) & (-32\%) \\
\end{tabular}
}
\end{table}
\vspace{-.6cm}

\section{Mitigation}
Several techniques for mitigating adversarial attacks have been suggested in the literature. In this section, we discuss these methods and their impact on Ulixes.

\noindent\textbf{Adversarial Training}
Perhaps the most common countermeasure to adversarial examples is adversarial training. As described in \cite{adv_training}, this technique consists in including adversarial examples in the training dataset of a classifier. Experiments from \cite{adv_training_success} show how this practice enhances the robustness of a model against adversarial examples and makes it harder to generate new ones.

\begin{figure}
\includegraphics [width=0.8\columnwidth] {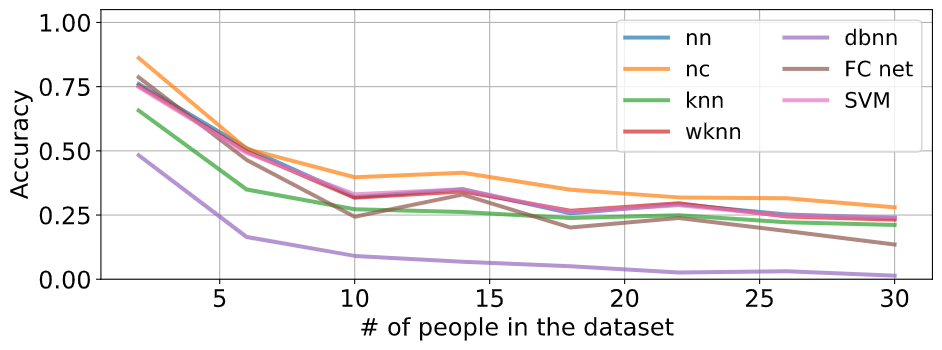}
\caption{\label{img:id_test_adv_training}
    Performance of classifiers trained on adversarial examples generated by Ulixes. The experiments are run for $15$ clear and $15$ cloaked images per identity leaked to the classifiers.
}
\end{figure}

We conduct a series of experiments to assess how adversarial training can impact the efficacy of Ulixes' cloaks. Referring back to the results in Fig. \ref{img:id_test}, we rerun those same experiments but leak cloaked images to the classifiers for training. Fig. \ref{img:id_test_adv_training} shows the results of such experiments when classifiers are trained on both original and cloaked images. 

On a similar note, an attacker could augment a training dataset by including small transformations to images, such as shifting or adding noise. These types of transformations were considered in \cite{Fawkes} and were found to have a negligible effect on the performance of this type of adversarial attack.

\noindent\textbf{Poison Attack and Anomaly Detection}
Another possible approach to mitigating generated cloaks is through poison attack detection \cite{poisoning} \cite{poisoning2}. These methods typically look for outliers in the dataset as indications of poisoning. However, our masks work by shifting the image in the feature space from the correct image cluster to an adjacent cluster. Thus the cloaked images are not outliers but appear to be legitimate members of an (incorrect) class. This is further proved by the measured distances between faces, as the maximum distance found between any pair of embeddings was $1.8$ and many pairs of embeddings had similar distances. Because FaceNet's Euclidean space allows a maximum distance of $4$, no samples can be said to have been pushed so far from the others to be considered outliers.

On the other hand, if an attacker has access to the original version of an image in the process of being classified, comparing the two's DSSIM score and L2 norm would show that while the images are similar their features are far apart, hinting at the presence of a cloak. If the images are assumed to be unlabeled, then knowledge of the presence of a cloak cannot be exploited in any way. However, if an attacker is confident about the label of one of the two images and this is also surely uncloaked, then the label of the other can be inferred easily. It must be noted that if an attacker already had a labeled image equal to a new one being submitted, then performing facial recognition with a neural network would be of little use: a simple visual comparison could achieve comparable results. Additionally, multiple versions of an image with different crops, compression levels, etc are common occurrences, making it difficult to ascribe the differences to cloaking.

\vspace{-.3cm}
\section{Conclusion \& Future Work}
Of the five target selection strategies proposed, the most effective strategy was selecting no target at all. This makes sense in the context of feature extraction. Neural network classifiers are more vulnerable to attacks that make them mislabel a sample for another particular sample, thus focusing on introducing disturbances in images towards a specific output class. Feature extractors, on the other hand, do not have any pre-defined output classes, and this research shows that non-targeted strategies, rather than directing an adversarial attack towards a particular target, yield optimal results.

While Ulixes outperforms both Fawkes and Face-Off, in both our transferability tests and versus an online API, the non-poisoned dataset case remains a challenging problem. There is significant room for improvement in future works, and given the prevalence of existing uncloaked images on social media, it is a practically relevant problem. Of particular concern is the poor transferability observed between models with substantially different architectures, since transferability is the key property required to apply these techniques to black-box algorithms. Our ensemble techniques are an initial step in solving this problem, but there is substantial room to improve these algorithms. Another area we plan to explore is the extension of our algorithm beyond the triplet loss function to other loss functions such as contrastive loss\cite{triplet_constrastive_loss_1}\cite{triplet_constrastive_loss_2}.
\section{Acknowledgments}
This research received no specific grant from any funding agency in the public, commercial, or not-for-profit sectors.

\bibliographystyle{IEEEtran}
\bibliography{main}

\end{document}